# Nonparametric Sparse Representation

Mahmoud Ramezani Mayiami
ECE Department, Shahed University
Tehran, Iran
ma.ramezani@shahed.ac.ir

Babak Seyfe
ECE Department, Shahed University
Tehran, Iran

*Abstract*—This paper suggests a nonparametric scheme to find the sparse solution of the underdetermined system of linear equations in the presence of unknown impulsive or non-Gaussian noise. This approach is robust against any variations of the noise model and its parameters. It is based on minimization of *rank pseudo norm* of the residual signal and $l_1 - norm$ of the signal of interest, simultaneously. We use the steepest descent method to find the sparse solution via an iterative algorithm. Simulation results show that our proposed method outperforms the existence methods like OMP, BP, Lasso, and BCS whenever the observation vector is contaminated with measurement or environmental non-Gaussian noise with unknown parameters. Furthermore, for low SNR condition, the proposed method has better performance in the presence of Gaussian noise.

## I. INTRODUCTION

THE approaches for over complete sparse decomposition aim at representing the signals as the combination of a few numbers of columns, namely atoms, which are chosen from an over complete dictionary. Recently, this problem has been applied in a number of fields, like Compressed Sensing or Compressive Sampling (CS) [1], [2] and Blind Source Separation (BSS) [3].

The main assumption in this framework is that the signal has a sparse representation (SR). It means that the total number of nonzero elements, $K$ in the signal coefficient vector is far less than the vector dimension $N$. Roughly speaking, the main problem is to find the sparsest possible $s \in \mathcal{R}^N$ from signal vector $r = \Phi s + n$ when $r$, $n$ are $M \times 1$ signal and noise vectors respectively and $\Phi$ is $M \times N$ dictionary where $M < N$. A standard formulation for the SR problem is given by

$$\hat{s} = \arg \min_s \|s\|_0 \; subject \; to \; \| r - \Phi s \|_2^2 \leq \delta. \quad (1)$$

where δ is the threshold of error and $\|.\|_0$ counts the nonzero elements. In general case, $\|s\|_p = \sqrt[p]{\sum_{i=1}^N |s_i|^p}$ for $p \geq 1$.

A number of approaches have been proposed to approximate $s$ based on (1) by using pursuit algorithms. For example, Mallat and Zhang [4] have used the maximum correlation between the residual signal $(r - \Phi \hat{s})$ and the atoms of the dictionary to find one of the nonzero elements in each iterations, namely Matching Pursuit (MP). Also, there are other iterative algorithms which used MP with some modifications to find the active coefficients of $s$ and their amplitudes, like Orthogonal MP (OMP) [5] and Stagewise Orthogonal MP (StOMP) [6].

Since finding the sparsest solution via (1) is an NP-hard combinatorial problem [7], another solution for reconstructing $s$ from $r$ is given by

$$\hat{s} = \arg \min_s \|s\|_1 \; subject \; to \; \| r - \Phi s \|_2^2 \leq \delta. \quad (2)$$

or equivalently

$$\hat{s} = \arg \min_s \| r - \Phi s \|_2^2 + \lambda \|s\|_1. \quad (3)$$

here $\lambda$ is the Lagrangian multiplier and it controls the effect of the sparsity terms $\|s\|_1$ in the solution. Basis Pursuit (BP) is one of the most familiar approach which solve (3) by using linear programming [8]. Moreover, researchers applied Bayesian framework to solve (3), like sparse Bayesian learning and the relative vector machine (RVM) [9], and Bayesian Compressive Sensing (BCS) [10].

In all of the conventional approaches, researchers have assumed that $n$ has Gaussian model. However, Middleton [11] showed that human activities in urban area may produce some noises whit non-Gaussian distribution. Furthermore, non-Gaussian model for noise were used in some applications [12]-[14]. In general, the parameters of the impulsive or non-Gaussian model of noise were supposed to be unknown. Moreover, the non-Gaussian noise for sparse representation problem were considered; for example, Wright and his colleagues [15], found the sparse solution in the presence of impulsive noise via the following optimization problem

$$\hat{s} = \arg \min_s (\|s\|_1 + \|n\|_1) \; subject \; to \; r = \Phi s + n. \quad (4)$$

Since they used some labeled training images for recognizing the class of the sparse images, the proposed approach in [15] was restricted to apply for object recognition and computer vision problem. Therefore, we cannot apply this method for solving SR problem in the presence of general non-Gaussian model for environmental or measurement noise while it arises in many situations in signal processing or communication area. More precisely, suppose a wireless communication noisy channel with non-Gaussian model of noise. The transmitter sends the sparse signal $s$ with encoded form $\Phi s$ through this channel and the receiver observes the signal $r = \Phi s + n$ (like compressive sensing measurement vector) while the model and the parameters of the noise vector $n$ are unknown for the receiver. The receiver aims to find the sparse coefficient vector $s$ without any knowledge but the elements of $r$ and $\Phi$. In this paper, we investigate SR problem in the presence of ambient impulsive noise which is non-Gaussian with unknown parameters.

Seyfe and Sharafat [16] used the properties of pseudo norm to design nonparametric multiuser detector that does not need any a priori information about the ambient noise model.

At the same way, in this paper, a new optimization algorithm for recovering sparse signals based on minimization of rank pseudo norm of $(r - \Phi s)$ is proposed while we minimize $l_1 - norm$ of the signal of interest $s$, simultaneously. Our proposed approach is robust in the sense of the variations in the noise model and parameters since we do not use any a priori knowledge about the ambient noise.

The reminder of this paper is organized as follows. Some preliminaries about pseudo norm, especially rank pseudo norm are defined in section II. In section III, our main idea is proposed to find the sparse solution in the presence of non-Gaussian noise. The section IV and V reserved for simulation and conclusion, respectively.

## II. CHARACTERISTIC OF PSEUDO NORM

As is well known, the norm function $\|.\|$, is defined on $\mathcal{R}^N$ and has four characteristics for $l$ and $m \in \mathcal{R}^N$ as following
(a) $\| l + m \| \leq \| l \| + \| m \|$,
(b) $\| \alpha l \| = |\alpha|. \| l \| \text{ for } \alpha \in \mathcal{R}$,
(c) $\| l \| \geq 0 \text{ for all } l \in \mathcal{R}^N$,
(d) $\| l \| = 0 \text{ if and only if } l = 0_N$. (5)

Where $0_N$ is a $N \times 1$ all zero components vector. The pseudo norm satisfies the first three properties of the norm function, but it has a looser constraint instead of (d). These properties are as follows
(a') $\| l + m \|_* \leq \| l \|_* + \| m \|_*$,
(b') $\| \alpha l \|_* = |\alpha|. \| l \|_* \text{ for } \alpha \in \mathcal{R}$,
(c') $\| l \|_* \geq 0 \text{ for all } l \in \mathcal{R}^N$,
(d') $\| l \|_* = 0 \text{ if ond only if } l_1 = l_2 \ldots = l_N$. (6)

where $\|.\|_*$ denotes to the pseudonorm operator and $l_i, i = 1,2, \ldots, N$, are the components of the vector $l$.

Now, we define the rank pseudo norm by

$$\| l \|_\varphi = \sum_{i=1}^{N} a(R(l_i))l_i. \quad (7)$$

Here $R(l_i)$ is the rank of $l_i, i = 1,2 \ldots, N$ among $N$ elements of the input vector $l$. Also, $a(.)$ is a score function which satisfies $a(1) \leq a(2) \ldots \leq a(N)$ and $\sum a(i) = 0$ [17]. It was proved that the rank pseudo norm with $a(i) = -a(N + 1 - i)$ is a kind of pseudo norm [16].

The general rank score was proposed [17] as $a_\varphi(i) = \varphi\left(\frac{i}{N+1}\right)$ whenever $\varphi(.)$ satisfies the following constraints

$$\int_0^1 \varphi(x)d(x) = 0, \int_0^1 \varphi^2(x)dx = 1. \quad (8)$$

The Wilcoxon function $\varphi(x) = \sqrt{12}(x - 0.5)$ and the sign function $\varphi(x) = sgn(2x - 1)$ are two kinds of score function $a(.)$ [16], [17].

## III. MAIN IDEA

In the majority of works in the SR research domain, the observation vector is contaminated with Gaussian noise. Hence, the optimum norm function which minimizes the noise power is Euclidean norm. Thus (1), (2) or (3) is an appropriate formulation for SR problem in the presence of known Gaussian noise. However, our simulations show that these optimization approaches fail when the sparse linear model is considered with unknown impulsive or non-Gaussian model of noise.

On the other hand, the characteristics of the rank pseudo norm cause this function to be an efficient tool for overcoming the effect of non-Gaussian noise in the sparse linear regression model [18]. Hence, we suggest our main optimization formulations as follows

$$\hat{s} = arg\min_{s} \|s\|_1 \, subject \, to \, \| r - \Phi s \|_\varphi \leq \zeta. \quad (9)$$

or equivalently

$$\hat{s} = \arg\min_{s} \| r - \Phi s \|_\varphi + \lambda \|s\|_1. \quad (10)$$

Since $\| r - \Phi s \|_\varphi$ is a continues and convex function of $s$ [19], we can find the sparse vector $\hat{s}$ through solving $\nabla f(s) = 0_N$, where $f(s) = \| r - \Phi s \|_\varphi + \lambda \|s\|_1$ as following

$$-s^T a(R(r - \Phi s)) + \lambda u(s) = 0. \quad (11)$$

where $(.)^T$ denotes to the vector transpose. Since $\|s\|_1$ is a non-differentiable function, we use the subgradient $u(s)$ instead of the gradient (see the Appendix). Also we have

$$a(R(r - \Phi s)) = \left(a\left(R(r_1 - \Phi_1^T s)\right), a\left(R(r_2 - \Phi_2^T s)\right) \ldots, a(R(r_M - \Phi_M^T s))\right)^T. \quad (12)$$

Here, $\Phi_i, i = 1,2 \ldots, M$ is the $i$-th row of the dictionary. There is not a closed form for the solution of (11), since both of its terms are nonlinear with respect to $s$. Therefore, we have investigated the iterative solution for recovering $s$ via (11). But we have achieved a sparse solution with relatively good performance compared with the commonplace reconstruction methods. Instead, we minimize the optimization problem (10) directly through the iterative algorithms. We applied the steepest descent method with using the $l_1 - norm$ to find the descent step direction $\Delta s_{sd}$ and the exact line search for calculating the step length $t$ in each iteration (see [20] for more details).

All things considered, our proposed algorithm for reconstructing $s$ from the optimization formulation (10) via steepest descent method is written in Algorithm 1. We named this algorithm Non Parametric Sparse Representation or NPSR.

In this algorithm, first of all, since we know that our solution is sparse, a zero vector is considered to initialize the sparse vector $s$. Then, in each iteration loops, one element will be added to the sparse estimated vector $\hat{s}$. It causes the $\| \hat{s} \|_1$ to be increased while the reconstruction error $\| r - \Phi s \|_\varphi$ is decreasing. This loop must be repeated unless the reconstruction error becomes less than the threshold $\xi$.

**ALGORITHM I: NONPARAMETRIC SPARSE REPRESENTATION (VIA STEEPEST DESCENT METHOD)**

**Given Parameter**: dictionary $\boldsymbol{\Phi}$, noisy signal $\boldsymbol{r}$, and error threshold $\xi$.
**Initialization**: Set the iteration number $k = 0$, and the initial coefficient sparse vector $\boldsymbol{s}^{(0)} = \boldsymbol{0}$.
**Main Loop**:
1. Calculate the step descent direction:
   - Set $\boldsymbol{\Delta s} = \boldsymbol{0}$,
   - Calculate the gradient vector $\boldsymbol{v} = \boldsymbol{\Phi}^T . a\left(R(\boldsymbol{r} - \boldsymbol{\Phi s}^{(k-1)})\right) - \lambda \boldsymbol{u}(\boldsymbol{s}^{(k-1)})$
   - Find the index $i$ of $\max abs(\boldsymbol{v})$ and set $\boldsymbol{\Delta s}(i) = sign(\boldsymbol{v}(i))$.
2. Find the step size $t$ via exact line search,
3. Update: $\boldsymbol{s}^{(k)} = \boldsymbol{s}^{(k-1)} + t\boldsymbol{\Delta s}$,
4. Check the stopping rule: if $\|\boldsymbol{r} - \boldsymbol{\Phi s}^{(k)}\|_\varphi + \lambda\|\boldsymbol{s}\|_1 > \xi$, increment $k$ by 1 and go to step1,

**Output**: Return $\boldsymbol{s}^{(k)}$.

## IV. SIMULATIONS

In this section, the performance of NPSR is studied and compared with the previous ones like OMP, BCS, BP, and Lasso [21]. The Matlab code for BCS is available online at http://www.ece.duke.edu/~shji/BCS.html. Also, we use *SparseLab* toolbox for simulating other methods, available from http://www.sparselab.stanford.edu.

The $N \times 1$ ($N = 1000$) signal $\boldsymbol{s}$ is generated with $K = 20$ nonzero elements drawn from an i.i.d Gaussian pdf with zero mean and unit variance. The positions of nonzero coefficients are chosen uniformly at random. The elements of the dictionary $\boldsymbol{\Phi}$ are i.i.d realization of a Gaussian pdf with zero mean and variance $\frac{1}{M}$ ($M = 300$). In each trial, a noise vector is generated and added to $\boldsymbol{\Phi s}$. The total number of trials is set to 1000 and the estimations of the signal through different methods for each trial are found. The reconstruction error is calculated by $E\left(\frac{\|\boldsymbol{s} - \hat{\boldsymbol{s}}_*\|_2}{\|\boldsymbol{s}\|_2}\right)$, where $*$ denotes to one of the reconstruction methods like the OMP, BCS, BP, Lasso, and NPSR.

The program is run for 10 different values of noise variance and the results for three kinds of noise are shown in Figs. 1, 2, and 3. Fig. 1 and Fig. 2 confirm that our proposed method for sparse representation has a better performance compared with the previous schemes. Furthermore, Fig. 3 shows that the proposed approach has a lower reconstruction error in the presence of the Gaussian noise for low SNR where $SNR = \frac{\|\boldsymbol{s}\|_2^2/N}{\|\boldsymbol{n}\|_2^2/M}$.

## V. CONCLUSION

In this paper, we proposed an optimization approach for finding the sparse solution of linear regression model $\boldsymbol{r} = \boldsymbol{\Phi s} + \boldsymbol{n}$ in the presence of non-Gaussian and impulsive noise. We have called this method as Non Parametric SR because we do not use any a priori information about noise parameters and model to calculate the sparse solution analytically.

Furthermore, we found that our proposed algorithm outperforms the existing greedy methods (like MP, OMP, and StOMP) and Bayesian approaches.

## APPENDIX

Here, we reviewed the concept of subgradient for easy reference (more details are available in [22] and [23]). Roughly speaking, the subgradient is a generalization of gradient for a convex non differentiable function.

Definition: let $f: \chi \to \mathcal{R}$ be a convex function where $\chi \subseteq \mathcal{R}^N$ is a convex set. The vector $\boldsymbol{g} \in \mathcal{R}^N$ is a subgradient of non differentiable function $f$ at point $\boldsymbol{x}' \in \chi$ if $\boldsymbol{g}$ satisfies the following condition

$$f(\boldsymbol{x}) \geq f(\boldsymbol{x}') + \boldsymbol{g}^T(\boldsymbol{x} - \boldsymbol{x}'), \forall \boldsymbol{x} \in \chi. \quad (13)$$

There can be more than one subgradient at $\boldsymbol{x}'$ and the set of all subgradients is named subdifferential at $\boldsymbol{x}'$ and is denoted by $\partial f(\boldsymbol{x})$. If $f$ is differentiable, $\partial f(\boldsymbol{x}) = \{\boldsymbol{g}\}$ where $\boldsymbol{g} = \nabla f(\boldsymbol{x})$ and $\{\boldsymbol{g}\}$ denotes to the set of subgradient vectors. The subgradient has scaling and addition properties which have given by (14) and (15), respectively

$$\partial(\alpha f(\boldsymbol{x})) = \alpha \partial f(\boldsymbol{x}), for \ \alpha > 0. \quad (14)$$
$$\partial(f_1(\boldsymbol{x}) + f_2(\boldsymbol{x})) = \partial f_1(\boldsymbol{x}) + \partial(f_2(\boldsymbol{x})). \quad (15)$$

As is well known, if $\boldsymbol{x}^*$ is the minimize of convex differentiable function $f$, $\nabla f(\boldsymbol{x}^*) = 0$ must be satisfied at the same way. For a convex non differentiable $f$, the optimality condition generalizes as following

$$f(\boldsymbol{x}^*) = \min_{\boldsymbol{x} \in \mathbb{R}^N} f(\boldsymbol{x}) \Leftrightarrow \boldsymbol{0}_N \in \partial(\boldsymbol{x}^*). \quad (16)$$

which follows from the definition of the subgradient

$$f(\boldsymbol{x}) = f(\boldsymbol{x}^*) + \boldsymbol{0}^T(\boldsymbol{x} - \boldsymbol{x}^*), \boldsymbol{x} \in \mathbb{R}^N. \quad (17)$$

For example, when the function of interest is $\|\boldsymbol{s}\|_1$, the subgradient is given by

$$\frac{\partial f(\boldsymbol{x})}{\partial \boldsymbol{x}} = \boldsymbol{u}(\boldsymbol{s}), u_i = \begin{cases} sgn(s_i), s_i \neq 0 \\ [-1, +1], s_i = 0 \end{cases}. \quad (18)$$

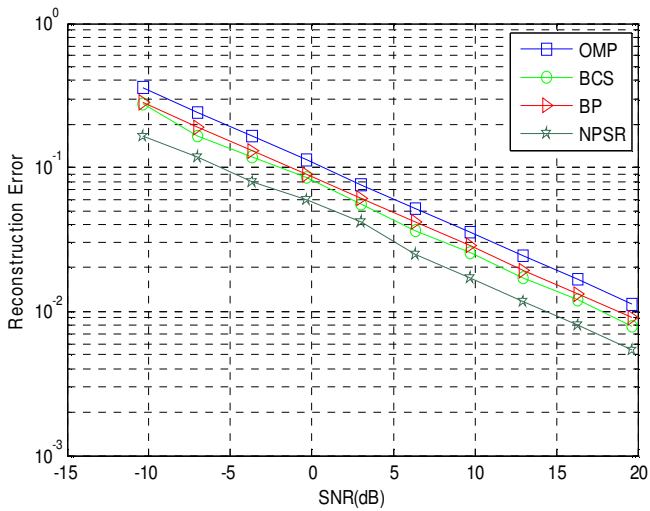

Fig. 1. NPSR versus $l_2$-norm based approaches for sparse representation in the presence of double-exponential noise.

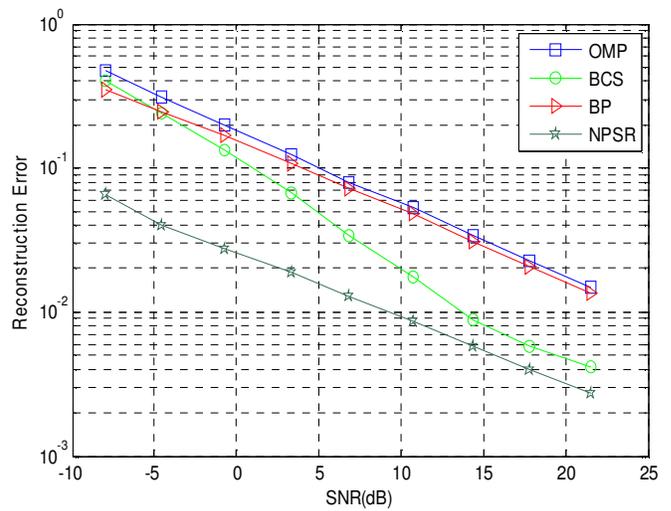

Fig. 2. NPSR versus $l_2$-norm based approaches for sparse representation in the presence of Middelton's class A model for non-Gaussian noise with $\kappa = 1000$ and $\varepsilon = 0.01$

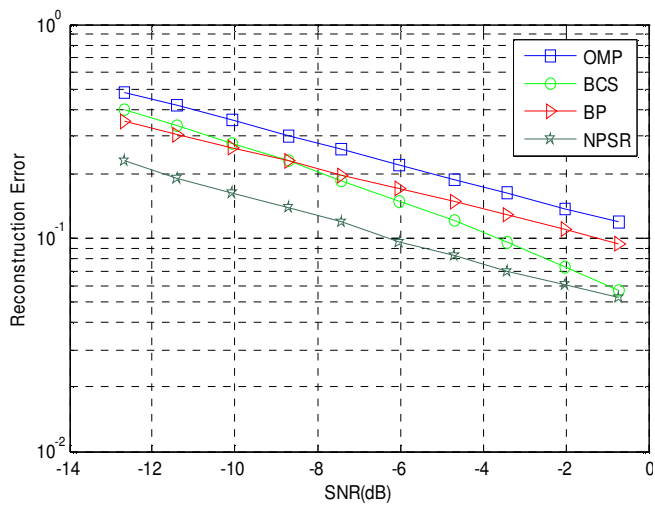

Fig. 3. NPSR versus $l_2$-norm based approaches for sparse representation in the presence of Gaussian noise.